\documentclass[11pt,a4paper]{article}

\usepackage{graphicx}

\usepackage[utf8]{inputenc}
\usepackage[T1]{fontenc}
\usepackage{lmodern}
\usepackage{geometry}
\usepackage{hyperref}

\usepackage{amsmath,amssymb}

\usepackage{titlesec}
\titleformat{\section}{\Large\bfseries}{\thesection}{1em}{}
\titleformat{\subsection}{\large\bfseries}{\thesubsection}{1em}{}

\newcommand{\mis}{\textsc{MIS}}
\newcommand{\model}{${\tt MISRO\ }$}
\newcommand{\mo}[1]{${\tt MISRO}_#1$}

\newcommand{\chuffed}{${\tt Chuffed}$}

\newcommand{\picatsat}{${\tt PicatSAT}$}

\newcommand{\highs}{${\tt HiGHS}$}

\newtheorem{problem}{Problem}
\newtheorem{definition}{Definition}

\usepackage{listings}
\usepackage{xcolor} 

\lstdefinelanguage{minizinc}{
	morekeywords={
		include, constraint, solve, maximize, array, of, var, int, float, set, forall, in, output, function, let, if, then, else, endif, case, return, assert, bool, tuple, ann, record
	},
	sensitive=true,
	morecomment=[l]{\%},
	morestring=[b]",
}

\lstdefinestyle{minizincstyle}{
    language=minizinc,
    basicstyle=\ttfamily\small,
    keywordstyle=\color{blue},
    commentstyle=\color{gray},
    stringstyle=\color{red},
    showstringspaces=false,
    breaklines=true,
    frame=tb, 
    captionpos=b 
}

\title{\textbf{Optimizing Ethical Risk Reduction for Medical Intelligent Systems with Constraint Programming}}

\author{
	Clotilde Brayé\textsuperscript{1,2,3}, Aurélien Bricout\textsuperscript{1}, Arnaud Gotlieb\textsuperscript{2}, Nadjib Lazaar\textsuperscript{3}, Quentin Vallet\textsuperscript{2} \\
	\\
	\textsuperscript{1}Enovacom, France, \texttt{\{prenom.nom\}@nehs-digital.com}, \url{https://www.enovacom.com/} \\
	\textsuperscript{2}Simula Research Laboratory, Norway, \texttt{arnaud@simula.no}, \url{https://www.simula.no} \\
	\textsuperscript{3}LISN, Université Paris-Saclay, \texttt{\{nom\}@lisn.fr}, \url{https://www.lisn.upsaclay.fr/}
}

\date{} 

\begin{document}
	
	\maketitle

\begin{abstract}
Medical Intelligent Systems (\mis) are increasingly integrated into healthcare workflows, offering significant benefits but also raising critical safety and ethical concerns. According to the European Union AI Act, most \mis\ will be classified as high-risk systems, requiring a formal risk management process to ensure compliance with the ethical requirements of trustworthy AI. In this context, we focus on risk reduction optimization problems, which aim to reduce risks with ethical considerations by finding the best balanced assignment of risk assessment values according to their coverage of trustworthy AI ethical requirements. We formalize this problem as a constrained optimization task and investigate three resolution paradigms: Mixed Integer Programming (MIP), Satisfiability (SAT), and Constraint Programming (CP).
Our contributions include the mathematical formulation of this optimization problem, its modeling with the Minizinc constraint modeling language, and a comparative experimental study that analyzes the performance, expressiveness, and scalability of each approach to solving. 
From the identified limits of the methodology, we draw some perspectives of this work regarding the integration of the Minizinc model into a complete trustworthy AI ethical risk management process for \mis.

\end{abstract}

\section{Introduction}

Medical intelligent systems (\mis) are artificial intelligence-based systems designed for the healthcare domain \cite{cohen2022intelligent}. These systems expanded into all areas of healthcare and are used in day-to-day clinical tasks. \mis\ have proven to be a major benefit for patients and caregivers by improving patient quality of life and information, personalized therapies, and also for medical personnel by improving diagnosis quality, facilitating delicate surgery operations, reducing administrative time-consuming tasks, etc. As \mis\ demonstrates significant advantages, it also involves risks that can have a dramatic impact on patients and medical staff. Then, one key challenge is to  evaluate these risks and mitigate them while considering ethical requirements (reqs). By adopting the risk-based {\it "AI ACT"} regulation which integrates several ethical reqs such as transparency, confidentiality, non-discrimination, accountability or human oversight \cite{european_commission_2021}, the European Commission has established an evidence-based layer for high-risk applications. Noticeably, high-risk AI applications must implement a risk management process that meets ethical reqs collection and associated risk reduction \cite{khinvasara2023risk}. Although risk management has long been a mandatory req. for so-called {\it medical devices} \cite{international_organization_for_standardization_medical_2019}, its relationship to the ethical reqs of AI is new and calls for novel methodologies \cite{vyhmeister2024tai,slattery2024ai,braye2024}.
Risk management is carried out by a college of experts (CoE) selected specifically for the design of a given \mis{} and representing all potential users and stakeholders. For each step of the development lifecycle \mis, the CoE is responsible for estimating whether the risks are acceptable and whether the evaluation of ethical reqs leaves \mis{} within acceptable limits \cite{cohen2022intelligent}. Nowadays, effective CoEs work by gathering consensus during periodic meetings 
but there is a lack of support tool for risk evaluation and reduction, especially when AI ethical reqs must be met \cite{european_commission_2021}. Finding the best balanced trade-off when it comes to minimizing risks while satisfying ethical reqs leads to solving a difficult optimization problem.

In this article, we formalize and address this problem with constraint solving and optimization methods. The problem consists in finding the optimal quantification of the risks identified for the \mis\ at each phase of the lifecycle, while ensuring the satisfaction of its AI ethical reqs. The contributions of this paper are threefold:
\begin{enumerate}
\item We propose for the first time a mathematical formulation of this optimization problem that is currently addressed manually by experts. Our formulation establishes \model ({\it \mis\ Risk ethical req. Optimization}) as a constraint optimization problem that finds an optimal assignment of risk quantification and allows experts to select risk mitigation measures with objectivity;
\item \model is modelled using a Minizinc model~\cite{minizinc} that allows us to solve this problem at each phase of the lifecycle by using diverse methods such as MIP, SAT or CP. Note that a different   \model problem is solved at each phase as the set of risks evolves accordingly. Our modeling is sufficiently versatile to accommodate many risk estimation procedures that entail nonlinear (bilinear, quadratic) constraints;
\item We perform a comparative experimental study to analyze the performance, expressiveness and scalability of each constraint optimization approaches. By using three different solvers, namely
\chuffed, \picatsat{} and \highs, our analysis allows us to draw some clear conclusions on which method to select, how scalable is the best approach, and how nonlinear constraints affect solver performances. We also identify the limits of our approach and propose future work to partially cope with them. 
\end{enumerate}

\section{State of the Art}\label{sec:sota}
Our work is at the crossroads of three areas: i) Healthcare domain; ii) Trustworthy AI; iii) AI software engineering. In the literature, we can find various risk management frameworks that are domain-specific for these three areas. For example, the authors in \cite{khinvasara2023risk} explore the challenges and best practices for a risk management process for medical devices to begin before the development phase. Their process follows the reqs of standard ISO 14971 on risk management for medical devices that do not include AI. 
Vyhmeister and Gastane propose TAI-PRM in \cite{vyhmeister2024tai}, for Trustworthy AI Project Risk Management, a methodology that introduces the concept of ethical risks. They define ethical risks as conditions resulting from a lack of consideration of trustworthy AI reqs. Their methodology allows for the use of metrics to assess ethical risks in the specific context of industry 5.0. The authors in \cite{haakman2021ai} express the need to include the risk management process in lifecycle system models. They review existing lifecycle models by adding a risk assessment step after the development phase. In these frameworks, automatization is not proposed. Currently, the automatization of the risk management process consists of developing tools that support risk management and allow reporting to be generated. 
MONARC, developed in \cite{mathey2018risk}, is a tool able to support risk analysis, automatically generate statistics figures with the user's input and generate risk management report. However, this tool do not consider ethical reqs in the risk management process. Other automation focuses on building AI risks databases so the CoE can use the database to identify risks \cite{slattery2024ai}. Ontology and taxonomy are created to structure these databases \cite{golpayegani2022airo,golpayegani2022towards}. 

\section{Formalization of \model{}}\label{sec:misro}

\begin{figure}
\centerline{\includegraphics[width=\textwidth]{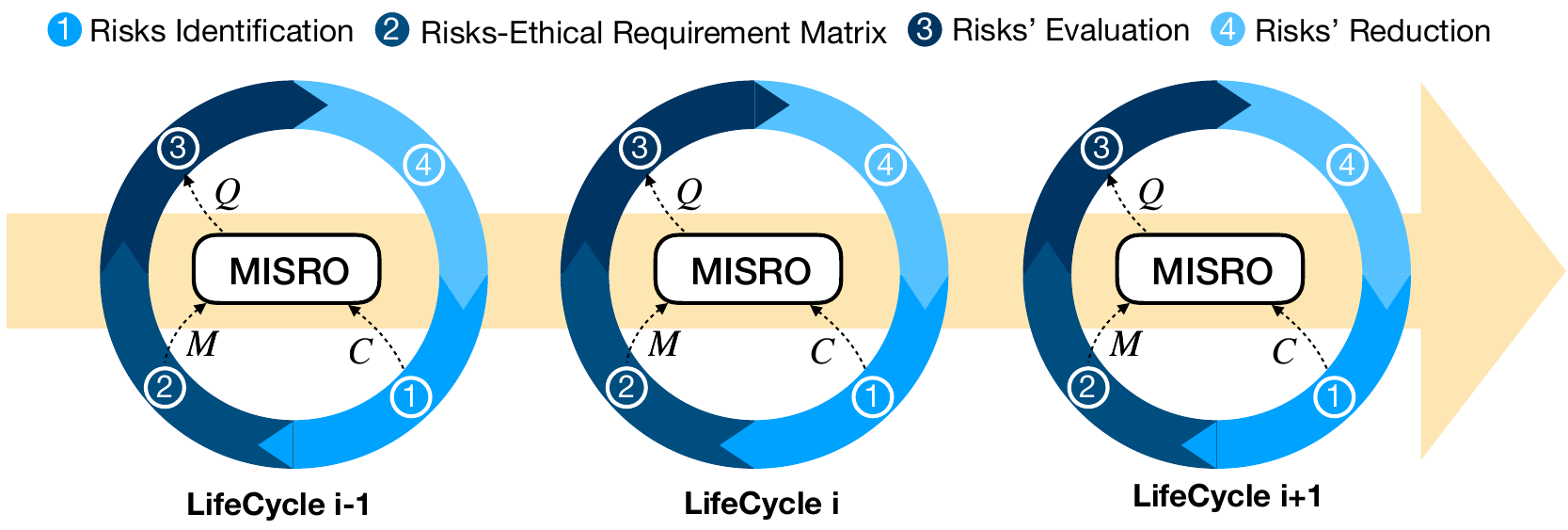}}
\caption{\mis\ Risk Management Process. Each lifecycle phase includes a risk evaluation cycle that updates risk quantification by solving the \model{}problem.}

\end{figure}

\subsection{Medical Intelligent System (\mis)}

A \mis\ is developed according to different lifecycle phases\cite{shearer2000crisp,amershi2019software}, such as req. collection, initial design, coding, testing, maintenance, etc. As \mis\ are commonly considered as high-risk AI systems, 
they must follow a risk management process. Hence, at each phase, risks for the \mis\ are identified, collected and evaluated. If necessary, a risk reduction step is undertaken \cite{muley2023risk}, so that the overall assessment shows that risk evaluation stays within acceptable limits. In parallel of this process, trustworthy AI ethical reqs have to be fulfilled for the \mis. So, into account for these considerations, we formalize \mis\ as follows:
\begin{definition}[Medical Intelligent System]
    Let $\mathbf{\mis}$ be a triple $(\mathcal{LC}, \mathcal{T}, \mathcal{R})$ such that: 
    \begin{itemize}
        \item $\mathcal{LC}=\langle lc_1, ..., lc_p\rangle$ is a finite ordered set representing the successive AI system's lifecycle phases ($p \geq 1$);
        \item $\mathcal{T}=\{t_1, \ldots, t_m\}$ is finite set of $m$ trustworthy AI ethical reqs ($m \geq 1$); 
        \item $\mathcal{R}=\{r_1, \ldots, r_n\}$ is a finite set of $n$ risks ($n \geq 1)$.\\ 
    \end{itemize}
\end{definition}

The \mis's lifecycle phases are ordered and executed sequentially, as shown in Fig.~\ref{fig:med_taid_process}. Starting from an evaluation of the reference criticality of the MIS (see Def.~\ref{def:rc}), the College of Experts (CoE) conducts an identification of the risks at each lifecycle phase. Using a so-called risks-ethical req. matrix (see Def.~\ref{def:rem}), an iterative process is launched which evaluates the risk quantification (see Def.~\ref{def:rq}) and applies mitigation measures until the risk criticality becomes acceptable w.r.t. the reference criticality.  

\subsection{Reference Criticality for a \mis}
Initially, for each \mis{} and for each trustworthy ethical req. of AI, the CoE defines a reference criticality level. This definition depends on various factors, such as the intended purpose of the system. For example, an automated skin cancer diagnostic tool would require a high level for 
for {\it explainability}, but a lower level for {\it environmental concerns}. To assist in assigning the reference criticality, the CoE could use existing methods, like assessing user explainability requirements \cite{10356408}.

\begin{definition}[Reference Criticality]\label{def:rc}
Given a \mis{} defined by the tuple $(\mathcal{LC}, \mathcal{T}, \mathcal{R})$, the \textbf{reference criticality} $C_{ref}$ is a vector 
\(
C_{ref} = [c(t_1),\ldots ,c(t_m)]^\top
\)
where $c$ is a function $c: \mathcal{T} \rightarrow [0, 1[$ that assigns to each ethical req. $t_i \in \mathcal{T}$ a real value representing its level of criticality. A lower value of $c(t_i)$ indicates that the req. $t_i$ is considered more critical and implying high expectations.
\end{definition}


For each phase of the lifecycle, reaching the reference criticality is the main objective of the CoE as it ensures control of the ethical risks associated with \mis. Note that the reference must be reached, as per regulation, at the end of each lifecycle step, including the evaluation phase, before deployment, and during post-production activations. 
\subsection{Risks Evaluation}
In the healthcare domain, each \mis{} can lead to potentially harmful situations affecting patients or medical personnel. These situations are referred to as \emph{risks}~\cite{international_organization_for_standardization_medical_2019}, 
and must be identified and systematically anticipated. 
All identified risks are collected in the set $\mathcal{R}$. 
For a given \mis, the CoE assigns to each risk $r_j \in \mathcal{R}$ two key attributes: likelihood of occurrence and severity level.

The \emph{likelihood} of a risk, denoted $l(r_j)$, represents the estimated probability that a harmful situation occurs within the context of the use of \mis. 
It takes values in the interval $[0, 1[$, where higher values indicate a higher risk. 
The value 1 is excluded, as this would imply a certain of occurrence, which contradicts the definition of a risk.
The \emph{severity} $s(r_j)$ quantifies the potential impact or severity of a harmful outcome should it occur. 
It is defined on $]0, 1]$, where higher values indicate more severe consequences. 
Since risks, by definition, involve harm, a severity level of zero is not possible.

The impact of the risk can then be determined considering both its likelihood and its severity. This process of combining these values to assess the general magnitude of a risk is known as {\it risk quantification}.

\begin{definition}[Risk Quantification]\label{def:rq}
    Let $r \in \mathcal{R}$ be a risk. The \textbf{risk quantification} is a mapping $q : \mathcal{R} \rightarrow [0,1[$ that assigns to each risk a value based on its likelihood and severity. 
\end{definition}
The exact mapping depends on the context and usually results from a CoE decision \cite{international_organization_for_standardization_medical_2019}. 
%
Several formulations of the risk quantification function $q$ have been proposed in the literature, each reflecting different assumptions about the interplay between likelihood and severity:
\[
\begin{array}{ll}
q_1(r) \triangleq \dfrac{l(r) + s(r)}{2} & \text{(arithmetic mean - linear)} \\
q_2(r) \triangleq l(r) \cdot s(r) & \text{(bilinear combination)} \\
q_3(r) \triangleq l(r) \cdot s^2(r) & \text{(quadratic emphasis on severity)}
\end{array}
\]
Each formulation results in a different distribution of risk scores, which influences how risks are prioritized. The arithmetic mean $q_1(r)$ treats likelihood and severity symmetrically, providing a balanced assessment. The linear combination $q_2(r)$ highlights risks where both likelihood and severity are simultaneously high, emphasizing joint significance. In contrast, the quadratic formulation $q_3(r)$ gives more weight to severity, capturing scenarios where the gravity of consequences is the dominant concern, even if the likelihood remains moderate. 

\subsection{Risks Reduction}

The quantified value of a risk can be reduced by taking a mitigation action. 
These actions are designed to reduce either the likelihood or the severity of a given risk, or both. 
These actions, proposed by the CoE, can take various forms, including the implementation of alert systems \cite{cozzo25}, anonymization procedures \cite{olat22}, as well as other context-specific interventions.


When a mitigation action is applied, it triggers a transition to the next phase of the risk management lifecycle.
This defines how a risk quantification evolves from one lifecycle phase ($lc_i$) to the next ($lc_{i+1}$), resulting in a lower value: $q'(r) < q(r)$. 
Mitigation actions are designed solely to reduce risk quantification.
However, no mitigation can fully eliminate a risk within a given phase and can introduce new risks that must be addressed in the next phase.

\begin{definition}[Mitigation Action]
Given a risk $r \in \mathcal{R}$ with quantification $q(r) \in ]0,1[$, a {\bf mitigation action} $a_r$ is a transformation that lowers the risk value:
\[
a_r : q \mapsto q' \quad \text{s.t.} \ q'(r) = q(r) - \varepsilon_r, \quad (\varepsilon_r > 0) \wedge (q'(r) \in ]0,1[).
\]
\end{definition}

\subsection{Mapping Risks to Ethical Requirements}
In order to relate the ethical reqs of trustworthy AI to risks associated with a given \mis, we have defined a matrix that quantifies the degree of correspondence between each risk and each ethical req.
Each entry in the matrix indicates the strength of the relationship between a specific risk and a particular req. 
The values range from 0 (no relation) to 1 (high similarity). 

\begin{definition}[Risk–Ethical Requirement Matrix]\label{def:rem}
    Given a set of ethical reqs $\{ t_i \}_{1 \leq i \leq m}$ and a set of risks $\{ r_j \}_{1 \leq j \leq n}$, we define the {\bf risk–ethical requirement matrix} $M = (m_{ij})$, an $m \times n$ matrix where each entry $m_{ij} \in [0,1]$ represents the degree of similarity between req. $t_i$ and risk $r_j$:
    \[
    M =
    \begin{pmatrix}
        m_{11} & m_{12} & \cdots & m_{1n} \\
        \vdots & \vdots & \ddots & \vdots \\
        m_{m1} & m_{m2} & \cdots & m_{mn}
    \end{pmatrix}
    \]
\end{definition}
This matrix, independent from the \mis, captures general relationships between ethical reqs and risks.

\subsection{Calculated Criticality}
To assess whether the risks associated with a \mis{} are acceptable, the CoE relies on a metric we define as \emph{calculated risk-based criticality}. 
This metric quantifies the extent to which each ethical req. of trustworthy AI is threatened by the risks currently being evaluated. 
It is calculated as a weighted average of risk quantifications, using the risk–ethical req. matrix as weights.
The calculated risk-based criticality is evaluated and compared to the reference criticality for each lifecycle phase. 
The purpose of risk mitigation and the progression of the lifecycle is to ensure that the calculated criticality does not exceed the reference criticality at the end of the process.

\begin{definition}[Calculated Risk-Based Criticality]\label{def:calc}
Let $(\mathcal{LC}, \mathcal{T}, \mathcal{R})$ be a \mis, with:
\begin{itemize}
  \item 
  $Q = [q(r_1), \ldots ,q(r_n)]^\top$ the vector of risk quantifications for all $r_j \in \mathcal{R}$,
  \item $M = (m_{ij})$ the $m \times n$ risk–ethical req. matrix,
  \item $\lambda_i = \sum_{j=1}^n m(t_i, r_j)$ the normalization coefficient for each ethical req. $t_i$, and $\lambda = [\lambda_1, \ldots, \lambda_m]^\top$ the corresponding vector.
\end{itemize}
The \textbf{calculated risk-based criticality} $C_Q$ is defined as:
\[
C_Q = \frac{1}{\lambda} M Q = 
\begin{pmatrix} 
\frac{\sum_{j=1}^n m(t_1, r_j) \cdot q(r_j)}{\sum_{j=1}^n m(t_1, r_j)} \\
\vdots \\
\frac{\sum_{j=1}^n m(t_m, r_j) \cdot q(r_j)}{\sum_{j=1}^n m(t_m, r_j)}
\end{pmatrix}
\]
\end{definition}
We opted to normalize using similarity metrics rather than the number of risks to better reflect the relative importance of each ethical req. 
Furthermore, if all risk quantifications reach their maximum value, the resulting calculated criticality will also reflect the highest possible threat level. Normalizing by the sum of similarity metrics guarantees that the criticality of each ethical req. takes its maximum value precisely when all associated risks have been fully quantified.

\subsection{Optimization Problem}
Up to this point, the CoE has relied on a {\it trial-and-error} approach. 
Since exact risk quantification values are unknown in advance, a naive strategy would be to apply mitigation actions to all risks to drive their quantifications to the lowest possible values. 
However, this approach is often impractical due to the cost or feasibility of implementing certain mitigation actions. 
Therefore, the CoE selects which risks to mitigate and estimates whether the resulting reduction is sufficient to meet ethical thresholds.
To support this decision process, we solve the following optim. problem: we seek a risk quantification that maximizes the minimum quantification value across all risks while ensuring that the calculated risk-based criticality remains below the reference criticality level.\\

\begin{problem}[\mis\ Risk-Ethical Requirement Optimization (\mo)]
Given an \mis\ $ (\mathcal{LC},\mathcal{T},\mathcal{R})$, a risk-ethical req. matrix $M = (m_{ij})$, and a reference criticality vector $C_{\text{ref}}$, the \textbf{\model} problem consists of finding a risk quantification vector $Q^*$ that satisfies:
\[
\begin{array}{ll}
    Q^* = \arg\max_Q \left(\min_j Q_j\right) \quad \text{subject to:} \\
    \quad C_{Q^*} = \dfrac{1}{\lambda} M Q^* \leq C_{\text{ref}}
\end{array}
\]
\end{problem}

\section{\model{}as a MiniZinc Model}\label{sec:zinc}

\model{}could be directly formulated using real-number computations (e.g., using floating-point representations), but this is not ideal for several reasons:
\begin{enumerate}
    \item Human-based assessment of the severity and likelihood of risks is easier using integer scales rather than real values in $[0,1)$;
    \item Estimating the effects of mitigation actions is more intuitive with integers;
    \item Floating-point arithmetic can introduce numerical instability, leading to incorrect deductions.
\end{enumerate}

For these reasons, we chose to implement the optimization problem using the MiniZinc\footnote{\url{www.minizinc.org}} modeling language that supports both solving and optimizing linear and nonlinear integer constraints. 
MiniZinc offers a high-level declarative syntax that enables modular and maintainable model development. Importantly, MiniZinc is solver-independent, i.e., the same model can be executed using various backends, including SAT, MIP, and CP solvers. 
This flexibility enables fair comparisons to be made across different solving paradigms.



%

Fig.~\ref{fig:minizinc} shows the implementation of \model\ in MiniZinc, including the ${\tt .mzn}$ file describing the model and a sample instance provided as a ${\tt .dzn}$ file.
The initial part of the model declares the variables and arrays of variables used in the model (${\tt lines 2-17}$).
Each risk ${\tt j}$
is associated with an auxiliary variable ${\tt Q[j]}$ (${\tt line 17}$) representing its quantification score, computed from the decision variables ${\tt l[j]}$ (likelihood, ${\tt line 15}$) and ${\tt s[j]}$ (severity, ${\tt line 16}$). 
${\tt M}$ encodes the the input risks-ethical reqs matrix ($\tt {line 12}$), where each entry ${\tt M[i,j]}$ indicates the similarities between the risk ${\tt j}$ and ethical req. ${\tt i}$. ${\tt C}$ corresponds to the reference criticality vector (${\tt line 13}$).
The parameter mode (${\tt line 6-7}$) is used to distinguish 
the three different formulations of risk quantification, presented above, which are encoded as constraints (${\tt line 19-23}$). Note the encoding of these constraints has an impact over the performance of the model and that is why, it is for instance preferable to encode $q_3$ as ${\tt l[j]*pow(s[j], 2)}$ instead of ${\tt l[j] * s[j] * s[j]}$ to avoid multi-occurrence of decision variables.
Following the recommendations in \cite{pascarella2021risk,hussain2017problems}, we use six different degrees in terms of risk's severity (${\tt max\_s=6}$) and  risk's likelihood (${\tt max\_s=6}$) to provide a sufficient degree of granularity for a meaningful interpretation. These values can be qualitatively mapped as follows: 
$1$ (ultra residual), $2$ (residual), $3$ (minor), $4$ (moderate), $5$ (major), and $6$ (extremely severe).
To ensure that the cumulative impact of all risks regarding 
each ethical req. remains below its respective threshold, the Reference threshold constraint is enforced (${\tt line 26-27}$).
Each normalization coefficient in ${\tt lambda}$ 
reflects the total similarity weight associated with req. ${\tt i}$.
Eventually, the optimization function is encoded as maximization of ${\tt minQ}$ by assigning each variable the greatest possible value in domain according to ${\tt l}$ and ${\tt s}$ input variables order (${\tt line 31-34}$).

\begin{figure}
    \centering

    {
\begin{lstlisting}[
        style=minizincstyle,
    %    caption={Unified MiniZinc model for the \mo{} problem with parameterized quantification mode.},
        label={lst:minizinc_model}
    ]
% misro.mzn
int: n; % number of risks
int: m; % number of ethical requirements
set of int: Risks = 1..n;
set of int: Ethics = 1..m;
int: mode; 
% 1: linear, 2: bilinear, 3: quadratic
int: max_l; % upper bound for likelihood
int: max_s; % upper bound for severity
int: max_Q; % upper bound for risk quant.

array[Ethics, Risks] of int: M;
array[Ethics] of int: C;
array[Ethics] of int: lambda;
array[Risks] of var 1..max_l: l;
array[Risks] of var 1..max_s: s;
array[Risks] of var 1..max_Q: Q;

% Risk quantification constraint
constraint forall(j in Risks) (
  (mode=1) -> (2 * Q[j] = l[j] + s[j]) /\
  (mode=2) -> (Q[j] = l[j] * s[j]) /\
  (mode=3) -> (Q[j] = l[j] * pow(s[j],2)));

% Reference threshold constraint
constraint  forall(i in Ethics)(
 sum(j in Risks)
  (M[i,j] * Q[j])<=lambda[i] * C[i]);

% Objective: maximize the minimum Q[j]
var 1..max_Q: minQ;
constraint forall(j in Risks) (Q[j]>=minQ);

solve :: int_search([l[j] | j in Risks] ++ [s[j] | j in Risks], input_order, indomain_max, complete) maximize minQ;
\end{lstlisting}
    }

    \vspace{0.3em}

    {
\begin{lstlisting}[
        style=minizincstyle,
      %  caption={Example instance file using mode 1.},
        label={lst:minizinc_data}
    ]
% instance.dzn 
n = 20; m = 12; max_l = 6; max_s = 6;
max_Q = 216; mode = 1;
C = [154, 93, 208, 109, 32, ..., 214, 28];
M = [| 4, 0, 6, 6, 0, 1, ..., 9, 7 |
       7, 9, 6, 8, 9, 3, ..., 0, 2 |
       ...
       6, 6, 2, 8, 4, 4, ..., 8, 4 |];
lambda =[sum(j in 1..n)(M[i,j])|i in 1..m];
\end{lstlisting}
    }

   \caption{MiniZinc implementation of the unified \mo model (${\tt misro.mzn}$).}
    \label{fig:minizinc}
\end{figure}

\section{Experimental Evaluation}\label{sec:experiments}

To evaluate the performance and scalability of our constraint-based model \mo, we generated synthetic instances. Each instance is characterized by three parameters: 
the number of risks $\alpha \in \{5, 10, 50, 100, 150, 200, 250, 300, 400, 500\}$, the number of ethical reqs $\beta \in \{4, 50, 100, 150, 200, 250, 300, 400\}$, and a version index $\gamma \in \{1, \dots, 10\}$ to mitigate the impact of randomness. These values were selected to cover a wide range of scenarios, from small to large-scale problems, in order to explore both general performance and scalability under increasing problem complexity.
Instances are named using the format ${\tt inst\_\alpha\_\beta\_\gamma}$. 
In each instance, the Risk–Ethical Requirement matrix ${\tt M}$ and criticality vector ${\tt C}$ are randomly generated. Matrix values are integers in $[0,10]$, approximating real-valued confidence scores with a precision of 10. 
Decision variables ${\tt l}$ and ${\tt s}$ have domains fixed to $[1,6]$ as said above.

Preliminary tests were conducted on six different solvers to identify the most effective solver from each of the three main solving paradigms. Based on these results, we selected the best-performing solver per category, i.e., those that led to the least number of time-out results:

\begin{itemize}
    \item \textbf{\chuffed} for  CP, a state-of-the-art solver based on lazy clause generation, well-suited for handling nonlinear constraints and combinatorial structures efficiently.
    
    \item \textbf{\highs} for MIP, a modern, high-performance solver optimized for large-scale LP/MIP problems, offering advanced presolving and branching techniques.
    
    \item \textbf{\picatsat} for SAT-based solving, which compiles MiniZinc models into propositional logic and solves them using SAT technology, performing best on linear and logic-intensive formulations.
\end{itemize}

We evaluated each of the 81 distinct instance configurations by generating 10 randomized versions per configuration, resulting in a total of 810 benchmark instances. 
Each instance was solved with all three selected solvers, and we report averaged results across the 10 variants. Experiments were run on a machine with a 12th Gen Intel Core i7-1255U (1.7 GHz, 10 cores, 12 threads) and 32 GB of RAM, using a timeout of 300 seconds per solver per instance. The full MiniZinc model and all benchmark data are publicly available at \texttt{https://anonymous.4open.science/r/MISRO-7F4F/}.




\subsection{Research Questions}

This study investigates the effectiveness of different constraint-solving paradigms in addressing the complexity of our proposed model \mo. To this end, we formulate the following research questions:

\begin{itemize}
    \item[\textbf{RQ1}] {\bf Solver Paradigms and Practical Efficiency —} How do different solving paradigms— CP,  MIP, and SAT — compare in terms of runtime performance, scalability, and their ability to handle the structural complexity of the \model{} formulation?
    \item[\textbf{RQ2}] {\bf Solution Quality under Time Constraints —} When a solver fails to prove optimality within the allotted timeout, how close are the returned solutions to the true optimum? Can we rely on these solutions as high-quality approximations in practice?
    \item[\textbf{RQ3}] {\bf Impact of nonlinearity —} To what extent do nonlinear risk quantification formulations (from \mo{1} to \mo{3}) affect solver efficiency and scalability? How robust are different solvers when handling bilinear and nonlinear constraints introduced in more expressive formulations?
\end{itemize}

\begin{figure}
    \centering
    \includegraphics[width=\columnwidth]{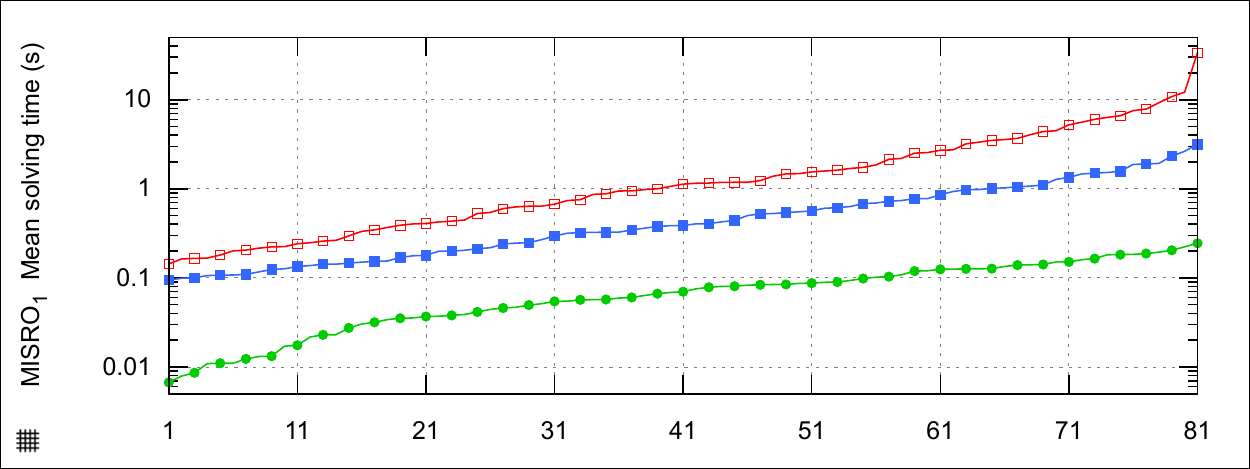}
    \includegraphics[width=\columnwidth]{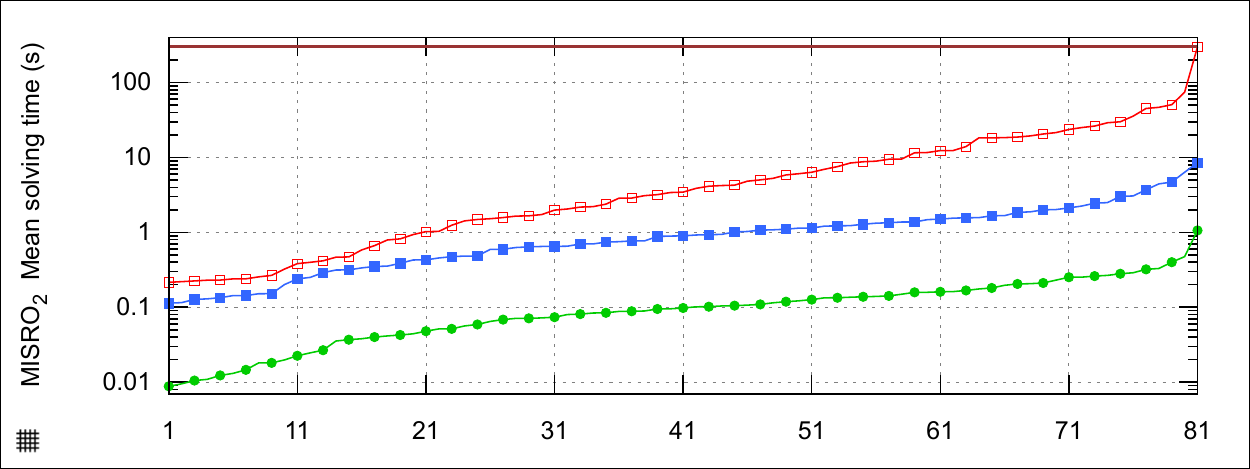}
    \includegraphics[width=\columnwidth]{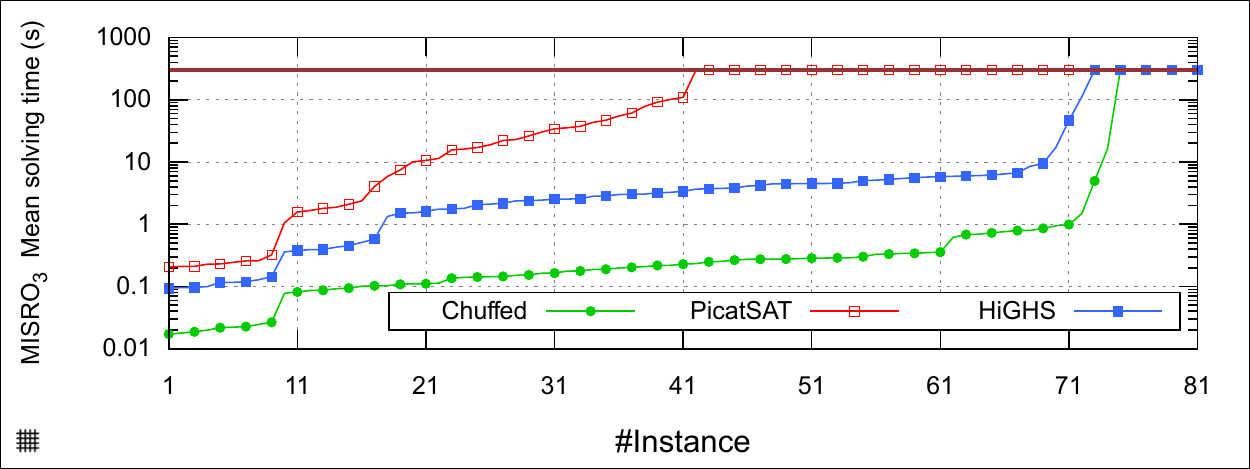}
\caption{Average CPU solving times (in sec) for the 81 instances per mode: \mo{1}, \mo{2}, and \mo{3}. Results for \chuffed{} (CP), \highs{} (MIP), and \picatsat{} (SAT).}
    \label{fig:time}
\end{figure}

\begin{figure}
    \centering
    \includegraphics[width=\columnwidth]{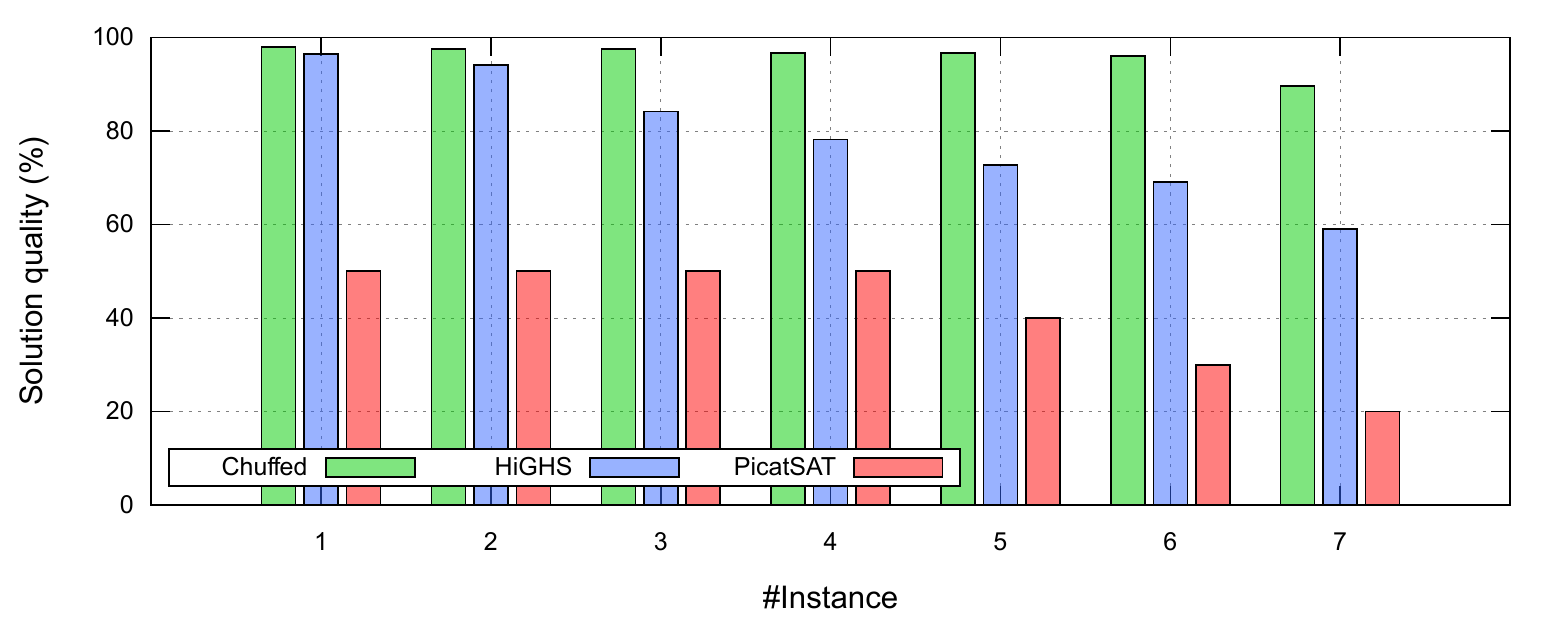}
  \caption{Solver performances on 7 timeout-prone hard instances.}
  \label{fig:rq2}
\end{figure}

\begin{figure}
    \centering
    \includegraphics[width=\columnwidth]{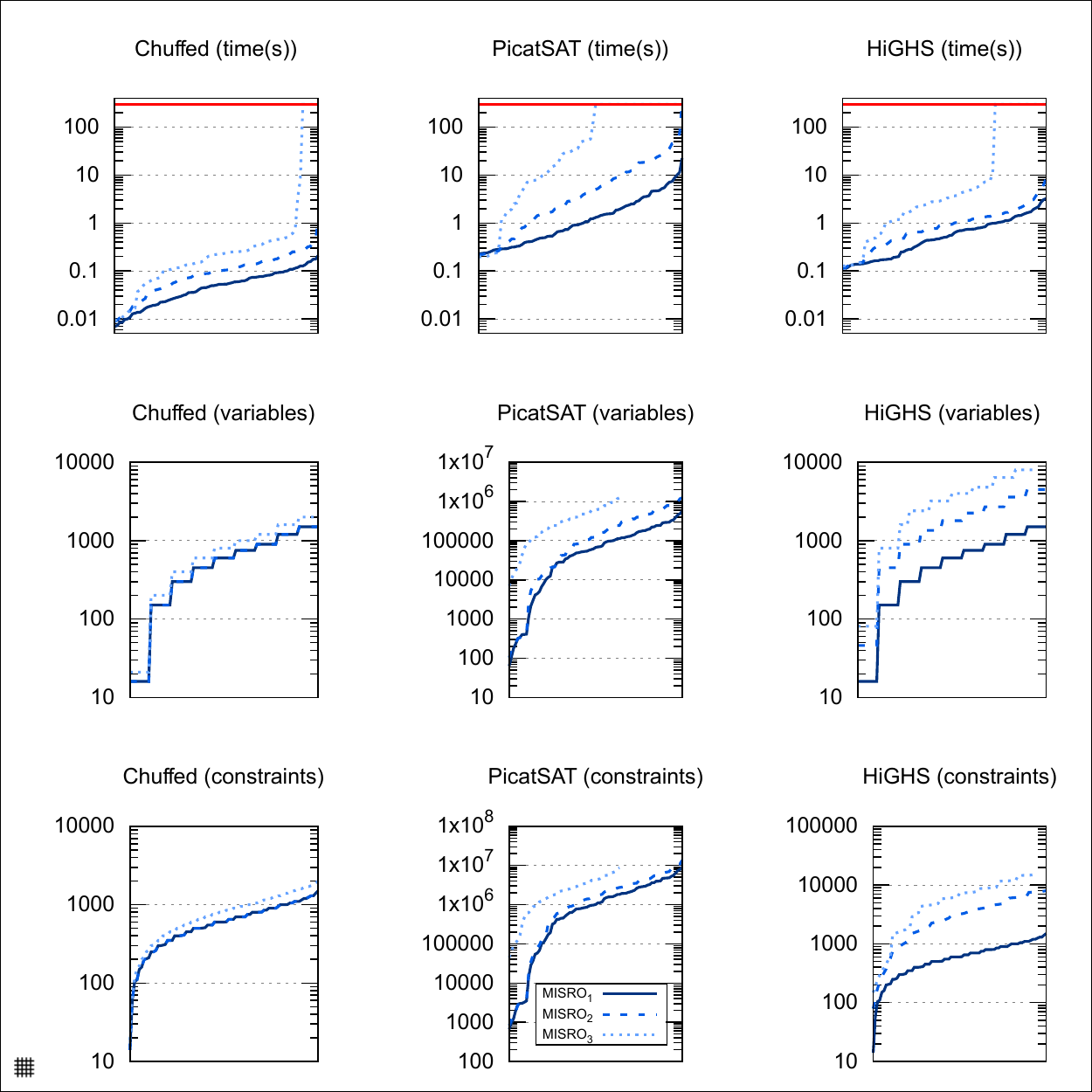}
  \caption{Solver performances comparison across three quantification functions.}
  \label{fig:rq3}
\end{figure}

\subsection{Results}

\subsubsection{RQ1 — Solver Paradigms and Practical Efficiency}

To assess the practical efficiency of different solving paradigms, we compared \chuffed{} (CP), \highs{} (MIP), and \picatsat{} (SAT) on all 81 instance configurations across the three risk quantification variants that we name \mo{1}, \mo{2}, and \mo{3}. 
Fig.~\ref{fig:time} reports the average CPU time to reach and prove optimality over 10 randomized variants per instance. 
Across all model variants, {\chuffed{} consistently outperforms the other solvers}, achieving the lowest average solving times and the highest robustness. Specifically, it solves each instance of \mo{1} in under 0.3 seconds, 
each one of \mo{2} in under 1 second, and most of \mo{3} within 4 seconds, with only 4 timeouts at the 300-second limit. 
These results highlight the strength of CP for this class of combinatorial optimization problems. The learning capabilities of \chuffed's lazy clause generation, combined with effective domain propagation, enable it to prune large parts of the search space efficiently, even in the presence of nonlinear constraints.

\highs, representing the MIP paradigm, shows competitive performance on \mo{1} (under 4 seconds) and \mo{2} (under 10 seconds), but struggles with the nonlinear structure of \mo{3}, where solving time increases significantly and 5 instances hit the timeout. Despite the required linearization, \highs{} benefits from strong LP relaxations and modern MIP heuristics.

By contrast, \picatsat{} performs well only on the simplest variant \mo{1}, with solving times exceeding 10 seconds. For \mo{2}, it regularly exceeds 100 seconds, and for \mo{3}, over 49\% of instances time out. The SAT-based approach suffers from the poor encoding efficiency of nonlinear arithmetic and the iterative optimization strategy it uses (e.g., branch-and-bound over successive SAT calls), which significantly increases solving overhead.

To sum up, CP is the most effective paradigm for solving \mo, especially when nonlinear quantification or user-defined constraints between risks are present. Its expressiveness and native support for complex constraints make it a strong candidate for both modeling flexibility and computational efficiency.

\subsubsection{RQ2: Solution Quality under Time Constraints}

When solvers fail to prove optimality within the time limit, a key concern is whether the returned solutions are still of acceptable quality. 
To assess this, we analysed the 7 instances for which none of the selected solvers—\chuffed{}, \highs{}, and \picatsat{}—were able to prove optimality within the 300-second timeout.

Fig.~\ref{fig:rq2} reports the solution quality, measured as the percentage ratio between the solver’s best solution and the known optimal value (i.e., $100\%$ corresponds to an optimal solution). Across all 7 instances.
\textbf{\chuffed{}} consistently returns high-quality solutions, with quality values ranging from \textbf{89.6\%} to \textbf{98\%}, confirming the robustness of CP-based search even under timeout.
\textbf{\highs{}} performs slightly below \chuffed{}, achieving solution qualities between \textbf{59\%} and \textbf{96.5\%}, showing its capacity to approach optimality in linearisable cases.
\textbf{\picatsat{}} returns notably weaker solutions, with all values fixed at either \textbf{50\%} or lower, due to the inefficiency of SAT-based solving in managing nonlinear arithmetic and optimization simultaneously.

These results suggest that, even under heavy time constraints, \chuffed{} and \highs{} can yield solutions that are close to optimal and thus practically useful. 
In contrast, \picatsat{} shows significant degradation in solution quality, highlighting its limitations in optimization-oriented scenarios. 
This reinforces the idea that the CP and MIP paradigms are more reliable for high-quality approximations under timeout.\\

\subsubsection{RQ3: Impact of nonlinearity}

To evaluate how increasingly nonlinear formulations affect solver performance, we compare the three versions of our model—\mo{1}, \mo{2}, and \mo{3}—under each selected solver. 
We recall that while \mo{1} uses a simple linear aggregation of risk parameters, \mo{2} introduces bilinear terms ($l_j \cdot s_j$), and \mo{3} further increases nonlinearity with quadratic terms ($l_j \cdot s_j^2$).

Fig.~\ref{fig:rq3} provides solver behaviour as model complexity increases. 
For each solver (\chuffed, \highs, and \picatsat), we report 
(i) the \textbf{average solving time (CPU)} required to reach or prove optimality in the 81 instances;
(ii) the \textbf{number of variables} and (iii) \textbf{number of constraints} generated after MiniZinc flattening, reflecting the size and structural complexity of the model compiled.

\subsection*{CPU Time Analysis}
\chuffed{} demonstrates strong robustness in all variants of the model. 
On \mo{1}, solving is near-instantaneous (average $<0.1$s) with all 81 instances solved optimally. 
For \mo{2}, which introduces bilinear terms, the solving times remain low (mainly $<0.3$ s), showing the efficiency of \chuffed in handling moderately complex arithmetic constraints. 
In \mo{3}, where quadratic risk modeling is added, the average time increases to approximately 8s.
Nevertheless, 77 out of 81 instances are solved within the 300s timeout. 
This performance is attributed to the support of \chuffed for non-linear arithmetic and its lazy clause generation engine, which uses conflict-driven learning to prune the search space effectively.

\highs{} performs well on \mo{1}, solving all instances with average times below 2s. 
The linear risk formulation aligns well with \highs's MIP solving capabilities. 
On \mo{2}, which includes bilinear terms, performance moderately degrades—most instances remain solvable within a few seconds, but some approach the timeout. 
For \mo{3}, with fully quadratic risk, \highs{} performs significantly worse: 21 instances reach the timeout and average times become skewed by large values. 
As a solver designed for linear and mixed-integer linear programs, \highs{} relies on linear approximations of nonlinear terms, which impairs its ability to scale with increased nonlinearity.

\picatsat{} is efficient on \mo{1}, solving nearly all instances under 1s. 
However, the introduction of bilinear terms in \mo{2} leads to larger CNF encodings and greater numbers of auxiliary variables, resulting in inconsistent performance and frequent timeouts. 
In \mo{3}, with nonlinear risk terms, 35 instances hit the 300s timeout. 
The main bottleneck arises from the iterative CNF encoding and flattening processes, which produce a large number of SAT calls and complex encodings. While SAT solving is fast in isolation, the overhead of handling nonlinear constructs makes this approach impractical for larger or more complex instances.

\subsection*{Model Growth Analysis}

With \chuffed{}, the number of variables remains constant between \mo{1} and \mo{2}, indicating that bilinear terms are handled internally without auxiliary variables. 
In \mo{3}, the number of variables increases by around 33\% (e.g., from 451 to 601), due to the inclusion of quadratic risk modeling. 
Constraints show similar growth: \mo{2} introduces minor increases, while \mo{3} significantly expands the constraint set to capture nonlinear logic.

For \highs, the size of the model increases dramatically between formulations. 
The number of variables triples from \mo{1} to \mo{2} (e.g., 151 to 451), and increases by an additional 70--80\% in \mo{3} (e.g., 451 to 801). 
Constraint counts scale similarly, with \mo{2} often doubling or tripling the number from \mo{1}, and \mo{3} sometimes doubling \mo{2}. 
This reflects the reliance of \highs on linear reformulations and auxiliary constructs to express non-linear constraints.

\picatsat{} experiences explosive growth in both variables and clauses. 
Variables grow from tens to hundreds in \mo{1} (e.g., 64), to over 550,000 in \mo{3}; clause counts follow suit, growing from tens of thousands to over 13 million. 
This reflects the exhaustive nature of the CNF transformation and the overhead incurred when encoding nonlinear constructs via Boolean logic. For many large \mo{3} instances, this growth makes model generation itself infeasible within the 300s limit, highlighting  the scalability limitations of SAT-based nonlinear optimisation.



\section{Threats To Validity}

Although our approach demonstrates the effectiveness of CP for reducing ethical risks in \mis{}, several limitations must be acknowledged:

\begin{itemize}
    \item \textbf{Arbitrary upper bound $M$.} 
    The model relies on a fixed, user-defined upper bound $M$ for risk quantification intervals. 
    This choice influences the domain size and, consequently, the solving performance. 
    More principled or data-driven approaches to setting this bound could improve robustness.
    
    \item \textbf{FD approximation of real-valued optimization.} 
    Our formulation encodes a problem over real-valued risk quantifications using finite domain (FD) constraints. 
    While this allows CP solvers to operate efficiently, the approximation introduces a semantic gap. 
    A formal correctness proof establishing the equivalence between the original real-valued problem and its FD encoding is currently lacking.
    
    \item \textbf{Nonlinear constraint isolation.} 
    Nonlinear constraints—particularly those used for quantifying risk—are fully integrated within the model. 
    However, their evaluation could be decoupled and precomputed or managed separately. 
    This would be especially beneficial in large-scale scenarios, where the number of risks and ethical reqs grows substantially.
    
    
    \item \textbf{Sensitivity to solver heuristics.} The performance of CP solvers is highly dependent on the choice of search and variable/value selection heuristics. 
    Our current model uses a fixed settings, but fine-tuning these heuristics may lead to significantly better solving times, particularly on challenging or large instances.
\end{itemize}
\section{Conclusion}\label{sec:conclusions}

This work demonstrates the applicability and effectiveness of CP for reducing ethical risk in \mis. 
Through a comparative analysis with MIP and SAT approaches, we demonstrate that CP not only achieves optimal solutions efficiently, but also offers strong expressiveness and flexibility. 
Beyond producing optimal risk quantifications, CP opens promising avenues for dynamic and interactive ethical risk management. 
CP can naturally accommodate evolving operational constraints. 
For example, if certain risk targets become unattainable, additional constraints, such as minimum quantification thresholds, can be introduced in real time by the CoE, enabling the solver to adapt and deliver new optimal solutions that remain consistent.

Future work will focus on integrating this optimization framework into a complete trustworthy AI risk management pipeline. 
This includes incorporating interactive feedback mechanisms and automating decision support for ethical oversight in accordance with the EU AI Act.

\bibliographystyle{abbrv}
\bibliography{references}

\end{document}